\documentclass[runningheads]{llncs}
\usepackage{microtype}
\usepackage{graphicx}
\usepackage{booktabs} 

\usepackage{hyperref}

\usepackage{subcaption}
\usepackage[]{microtype}
\UseMicrotypeSet[protrusion]{basicmath} 
\usepackage{lmodern}
\usepackage[T1]{fontenc}
\usepackage[utf8]{inputenc}
\usepackage{textcomp} 
\usepackage{mdframed}
\usepackage{import}
\usepackage{fancyvrb}
\usepackage{listings}
\usepackage{import}
\usepackage{wrapfig}
\usepackage{nicefrac}
\usepackage{ragged2e}
\usepackage{caption}
\usepackage{mystyle}
\usepackage{stackengine}
\usepackage{scalerel}
\usepackage{bbm}
\usepackage{algorithm}
\usepackage{algpseudocode}
\usepackage{eqparbox,array}
\usepackage{enumitem}
\usepackage{pgf}
\usepackage{transparent}
\usepackage{xcolor}
\usepackage{todonotes}

\usepackage{amsmath}
\usepackage{amssymb}
\usepackage{mathtools}

\usepackage{amsmath,amsfonts,bm}

\def\eqref#1{equation~\ref{#1}}

\def\1{\bm{1}}

\DeclareMathAlphabet{\mathsfit}{\encodingdefault}{\sfdefault}{m}{sl}
\SetMathAlphabet{\mathsfit}{bold}{\encodingdefault}{\sfdefault}{bx}{n}

\newcommand{\E}{\mathop{\mathbb{E}}}

\DeclareMathOperator*{\argmax}{arg\,max}

\DeclareMathOperator*{\LSE}{\text{LSE}}

\definecolor{blueTileColor}{HTML}{afafff}
\newcommand{\blueTile}{{\color{blueTileColor}\blacksquare}}
\newcommand{\yellowTile}{{\color{yellow}\blacksquare}}
\definecolor{brownTileColor}{HTML}{f4a460} 
\newcommand{\brownTile}{{\color{brownTileColor}\blacksquare}}
\definecolor{redTileColor}{HTML}{ff8b8b} 
\newcommand{\redTile}{{\color{redTileColor}\blacksquare}}

\newcommand{\temperature}{T}

\newcommand{\RV}[1]{{\mathcal{#1}}}  
\newcommand{\pival}{\mathbb{V}}
\newcommand{\concat}{\centerdot}
\newcommand{\RepresentationClass}{\Phi}
\newcommand{\encode}{\text{encode}}
\newcommand{\prefix}{\rho}
\newcommand{\suffix}{\sigma}
\newcommand{\Nodes}{N}
\newcommand{\Edges}{E}
\newcommand{\prefixtree}{\mathcal{T}}
\newcommand{\surprise}{h}
\newcommand{\psat}{\text{psat}}
\newcommand{\last}{\text{last}}
\newcommand{\size}{\text{size}}
\newcommand{\ex}{x}
\newcommand{\EX}{X}
\newcommand{\demo}{\trc^*}
\newcommand{\Demos}{\demo_1, \ldots, \demo_m}

\newcommand{\trc}{\xi}
\newcommand{\Path}{\text{Path}}
\newcommand{\eoe}{\$}
\newcommand{\envDist}{\delta}
\newcommand{\policy}{\pi}

\newcommand{\identify}{\mathcal{I}}

\newcommand{\spec}{\varphi}
\newcommand{\ConceptClass}{C}
\newcommand{\concept}{c}

\newcommand{\eqdef}{\mathrel{\stackrel{\makebox[0pt]{\mbox{\normalfont\tiny def}}}{=}}}
\newcommand{\Reals}{\mathbb{R}}

\newcommand{\Nat}{\mathbb{N}}

\renewcommand{\eqref}[1]{(\ref{eq:#1})}

\usepackage[capitalize,noabbrev]{cleveref}

\begin{document}
\title{Demonstration Informed Specification Search}
%
%
\author{
Marcell Vazquez-Chanlatte\and
Ameesh Shah\and
Gil Lederman\and
Sanjit A. Seshia
}
\authorrunning{M. Vazquez-Chanlatte et al.}
%
\institute{University of California, Berkeley, Berkeley CA 94720, USA}
\maketitle              
\begin{abstract}
  This paper considers the problem of learning temporal task
  specifications, e.g. automata and temporal logic, from expert
  demonstrations.
  Three features make learning temporal task specifications
  difficult: (1) the (countably) infinite number of tasks under
  consideration, (2) an a-priori ignorance of what 
  memory is needed to encode the task, and
 (3) the lack of gradients to guide the search for explanatory tasks.
  To overcome these hurdles, we propose \emph{Demonstration Informed
    Specification Search (DISS)}: a family of algorithms requiring only
  \emph{black box} access to (i) a maximum entropy planner and (ii) a
  task sampler from labeled examples.  DISS works by alternating
  between (i) conjecturing labeled examples to make the demonstrations
  more likely and (ii) sampling tasks consistent with
  conjectured labeled examples.
  In the context of tasks described by Deterministic Finite Automata,
  we provide a concrete implementation of DISS that efficiently
  identifies a task from only one or two expert demonstrations.

\keywords{Learning from Demonstrations \and Specification Mining}
\end{abstract}

\section{Introduction}
Expert demonstrations provide an accessible and expressive means to informally
specify a task, particularly in the context of human-robot
interaction~\cite{abbeel2004apprenticeship,DBLP:conf/icml/NgR00,ziebart2008maximum}.
In this work, we study the problem of inferring, from demonstrations, tasks
represented by formal \emph{task specifications}, e.g., automata and temporal
logic. The study of task specifications is motivated by their ability to (i)
encode historical dependencies, (ii) incrementally refine the task via
composition, and (iii) be semantically robust to changes in the workspace. We
ground and motivate this problem with an example.

\subsection{Motivating Example}\label{robottaskspec}
\begin{figure}[h]
\begin{subfigure}[t]{0.3\textwidth}
  \centering \scalebox{0.92}{ 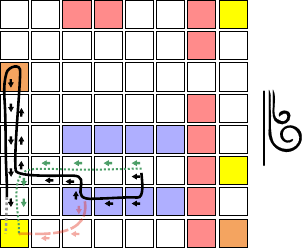 }
  \caption{\label{fig:gw8x8}}
\end{subfigure}
\hfill
\begin{subfigure}[t]{0.58\textwidth}
\centering
  \scalebox{.4}{ 
\begingroup%
  \makeatletter%
  \providecommand\color[2][]{%
    \errmessage{(Inkscape) Color is used for the text in Inkscape, but the package 'color.sty' is not loaded}%
    \renewcommand\color[2][]{}%
  }%
  \providecommand\transparent[1]{%
    \errmessage{(Inkscape) Transparency is used (non-zero) for the text in Inkscape, but the package 'transparent.sty' is not loaded}%
    \renewcommand\transparent[1]{}%
  }%
  \providecommand\rotatebox[2]{#2}%
  \newcommand*\fsize{\dimexpr\f@size pt\relax}%
  \newcommand*\lineheight[1]{\fontsize{\fsize}{#1\fsize}\selectfont}%
  \ifx\svgwidth\undefined%
    \setlength{\unitlength}{488.47432709bp}%
    \ifx\svgscale\undefined%
      \relax%
    \else%
      \setlength{\unitlength}{\unitlength * \real{\svgscale}}%
    \fi%
  \else%
    \setlength{\unitlength}{\svgwidth}%
  \fi%
  \global\let\svgwidth\undefined%
  \global\let\svgscale\undefined%
  \makeatother%
  \begin{picture}(1,0.47297901)%
    \lineheight{1}%
    \setlength\tabcolsep{0pt}%
    \put(0,0){\includegraphics[width=\unitlength,page=1]{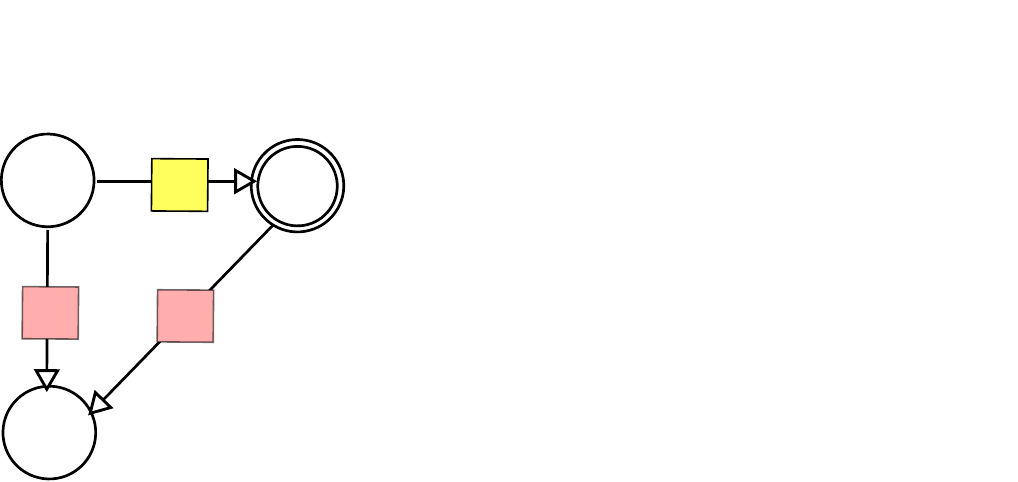}}%
    \put(0.04670961,0.29045387){\makebox(0,0)[t]{\lineheight{1.25}\smash{\begin{tabular}[t]{c}start\end{tabular}}}}%
    \put(0.70917231,0.29016491){\makebox(0,0)[t]{\lineheight{1.25}\smash{\begin{tabular}[t]{c}start\end{tabular}}}}%
    \put(0,0){\includegraphics[width=\unitlength,page=2]{true_dfa.pdf}}%
  \end{picture}%
\endgroup%
 }
  \caption{\label{fig:target-dfa}}
  \end{subfigure}
    \caption{DFAs with stuttering semantics, i.e., if a transition is not
    provided, a self loop is assumed. The accepting states are marked with a
    concentric circle and the initial state is labeled start. A sequence of
    inputs, e.g. colored, is accepted if the final state is accepting.  The
    left DFA encodes: ``Avoid $\redTile$ (lava) and eventually reach
    $\yellowTile$ (recharge)''.  The right DFA adds to the left DFA the rule:
    ``Visitations of $\blueTile$ (water) and $\yellowTile$ (recharge) must be
    separated by a visit to $\brownTile$ (drying).''}
\end{figure}

Consider an agent operating in the 8x8 grid world as shown in
Fig~\ref{fig:gw8x8}. The agent can attempt to move up, down, left, or
right. With probability \nicefrac{1}{32}, wind will push the agent
down, regardless of the agent's action. The black path is the
\emph{prefix} of an episode, in which the agent attempts to move left,
slips into the blue (water) tile~($\blueTile$), visits a brown
(drying) tile~($\brownTile$), and then proceeds downward.

Given the black demonstration, call $\trc_b$, and the \emph{prior} knowledge
that the agent's task implies that it will (i) avoid red (lava)
tiles~($\redTile$) and (ii) try to reach a yellow (recharge)
tile~$(\yellowTile)$, what task, as a Deterministic Finite Automaton (DFA),
explains the agent's behavior?

Upon inspecting the demonstration, one might be surprised that the agent goes
out of its way to visit $\brownTile$.  For example, why would the agent not
take the red dashed path directly to $\yellowTile$?  One might conjecture that
the agent's true task requires visiting $\brownTile$ after visiting
$\blueTile$. Similarly, one may notice that the demonstration is ``more
rational'' or ``less surprising'' if the dotted extension of $\trc_b$ ending in
$\yellowTile$ is a positive example of the task. Finally, appealing to Occam's
razor, one might search for a simple DFA consistent with the conjectured
labels. In this case, the demonstrated task is in fact one of the ``simplest''
DFAs, shown on the right in Fig~\ref{fig:target-dfa}.  We shall later
systematize this line of reasoning and provide a learner that recovers an
explanatory DFA similar to ground truth.

\subsection{Contributions}

This work introduces a family of approximate algorithms called
\emph{Demonstration Informed Specification Search} (DISS) that efficiently searches for
task specifications that explain a set of expert demonstrations. DISS leverages the following contributions:
\begin{enumerate}

\vspace{-0.6em}
    \item A proxy function whose gradient (i) informs the search
      for an explanatory task specification and (ii) is computed with
      \emph{black-box} access to a maximum entropy planner.
    \item A reduction from learning task specifications from demonstrations to learning from labeled examples.
    
\vspace{-0.4em}
\end{enumerate}
The resulting algorithm is \emph{agnostic} to the underlying concept class and dynamics. 
Finally, we provide a concrete implementation of
DISS. 
An example identification algorithm for DFAs and a maximum entropy
planner for simple gridworlds are also included. Using this
implementation, we perform two experiments validating that DISS indeed
enables efficiently searching for tasks that explain the
demonstrations even in large/unstructured concept classes like DFAs
given unlabeled and potentially incomplete demonstrations.

\begin{remark}
  The choice of DFAs as the concept class for our experiments was
  motivated by three observations. First, DFAs explicitly encode
  memory, making the contribution of identifying relevant memory more
  clear. Next, to our knowledge, all other techniques for learning
  finite path properties from demonstrations focus on syntax defined
  concept classes.  Thus, learning DFAs is understudied in this
  context. Third, DFAs constitute a very large and mostly unstructured
  concept class enabling studying the efficiency of
  DISS without introducing
  too many inductive biases. 
\end{remark}

\subsection{Algorithm Overview}
DISS assumes (i) access to a multi-set of expert demonstrations:
$\Demos$, (ii) \emph{black box} access to an identification algorithm,
$\identify$, that maps positively/negatively labeled paths to a
distribution over concepts and (iii) \emph{black box} access to a
planner that estimates the probability of a path given a candidate
task (see Sec~\ref{sec:agent_model}). DISS operates by cycling between
three components (shown in Fig~\ref{fig:overview}):
\begin{figure}[h]
  \centering \scalebox{0.9}{ 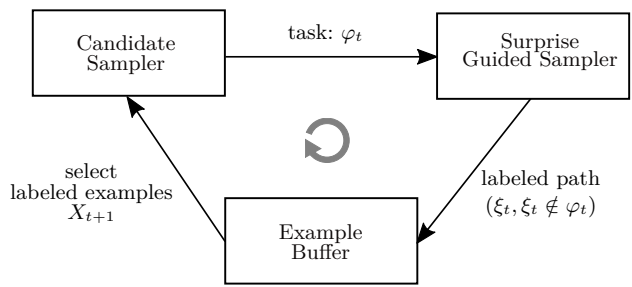 }

  \caption{Demonstration informed specification search overview.}
  \label{fig:overview}
\end{figure}

\begin{enumerate}
\item \textbf{Candidate Sampler}: A candidate task, $\spec$, is
  sampled from $\identify(\EX)$, where $\EX$ is a collection of
  labeled examples (initialized to the empty set).
\item \textbf{Surprise Guided Sampler}: The planner is used to find a
  path, where toggling whether the path is a positive/negative example
  of the task makes the expert demonstrations less surprising.
\item \textbf{Example Buffer}: Given previously seen data, the example
  buffer yields a set of positive and negative example paths.
\end{enumerate}
\subsection{Reading Guide}
The rest of the paper is structured as follows. We begin by highlighting
related work in Sec~\ref{sec:related-work} and formalizing our learning from
demonstrations (LFD) problem in Sec~\ref{sec:setup}. Next in
Sec~\ref{sec:agent_model}, we motivate employing an entropy regualized agent
model in our LFD problem. Groundwork laid, we discuss how to find surprising
paths given our agent model (Sec~\ref{sec:sggs}). This then leads us to propose
a variant of simulated annealing (implemented through the example buffer) to
approximately solve our LFD problem (Sec~\ref{sec:DISS}). Finally, we conclude
with an empirical study of the algorithm (Sec~\ref{sec:experiment}).

\subsection{Related Work}\label{sec:related-work}
The problem of learning objectives by observing an expert has a rich and well
developed literature dating back to early work on Inverse Optimal
Control~\cite{kalman1964linear} and more recently via Inverse Reinforcement
Learning (IRL)~\cite{DBLP:conf/icml/NgR00}. In IRL, an expert demonstrator
optimizes an \emph{unknown} reward function by acting in a stochastic
environment. The goal of IRL is to find a reward function that explains the
agent's behavior.  A fruitful approach has been to cast IRL as a Bayesian
inference problem to predict the most probable reward
function~\cite{ramachandran2007bayesian}. To make this inference robust to
demonstration/modeling noise, one commonly appeals to the principle of maximum
causal entropy~\cite{jaynes1957information,mce}. Intuitively, maximum causal
entropy models hedge and assign no more probability mass to an action than the
known statistics require, e.g., to match known expected speeds or location
visitation frequencies. In doing so, the maximum causal entropy models bound
the worst-case expected description length of future demonstrations~\cite{mce}.

While powerful, traditional IRL provides no principled mechanism for composing
the resulting reward artifacts and requires the relevant historical features
(memory) to be a-priori known. Furthermore, it has been observed that small
changes in the workspace, e.g., moving a goal location or perturbing transition
probabilities, can change the task encoded by a fixed
reward~\cite{DBLP:conf/nips/Vazquez-Chanlatte18,DBLP:journals/nips/Abel22}.

To address these
deficits, recent works have proposed learning Boolean task
specifications, e.g. logic or automata, which admit well defined
compositions, explicitly encode temporal constraints, and have workspace
independent semantics. The development of this literature mirrors the 
historical path taken in
reward based research, with works adapting optimal
control~\cite{DBLP:conf/cdc/KasenbergS17,DBLP:conf/rss/ChouOB20},
Bayesian~\cite{DBLP:conf/nips/ShahKSL18,DBLP:conf/icra/Yoon021}, and
maximum entropy~\cite{DBLP:conf/cav/Vazquez-Chanlatte20}
IRL approaches.

A key difficulty for the task specification inference from
demonstrations literature is how to search an intractably large (often
infinite) concept class. In particular, and in contrast to the reward
setting, the discrete nature of automata and logic, combined with the
assumed \emph{a-priori} ignorance of the relevant memory required to
describe the task, makes existing gradient based approaches either
intractable or inapplicable. Instead, the majority of current literature either
(syntactically) enumerates
concepts~\cite{DBLP:conf/nips/Vazquez-Chanlatte18,DBLP:conf/nips/ShahKSL18,DBLP:conf/icra/Yoon021}
or hill climbs via simple probabilistic (syntactic)
mutations~\cite{DBLP:conf/cdc/KasenbergS17,DBLP:phd/basesearch/Carrillo21}.

The major contribution of this paper is to systematically reduce the
problem of learning from demonstrations into a series of supervised
task specification identification problems, e.g., finding a small
DFA that is consistent with a set of
example strings~\cite{DBLP:conf/icgi/HeuleV10}, a problem more
generally referred to as Grammatical
Inference~\cite{de2010grammatical}.  The result is a principled way to
sample tasks given a candidate task.  This insight is integrated into
a variant of simulated annealing~\cite{DBLP:conf/wsc/SkiscimG83} for
\emph{guided} hill climbing. Further, if one assumes that the
demonstrations are \emph{positive examples}, our work serves as a
computationally tractable approach to learning simple regular
languages from simple positive
examples~\cite{DBLP:journals/ml/Denis01} in the context of Markov
Decision Processes.

Finally, we note some similarity with the approach
of~\cite{DBLP:conf/rss/ChouOB20} which uses an approximate optimal 
control perspective to hypothesize positive and negative
examples to constrain its search when given demonstrations in
a deterministic domain. This work strictly generalizes this
setting while providing (i) a belief over a distribution of tasks
(ii) being agnostic to the concept class, (iii) supporting
stochastic domains, (iv) supporting backtracking, and (v) not assuming
(approximate) optimal performance from the demonstrations.

\section{Preliminaries and Problem Statement}\label{sec:setup}

\subsection{Dynamics Model} We model the expert \emph{demonstrator} as
operating in a \emph{Markov Decision Process} (MDP), $M = (S, A, s_0,
\envDist)$, where (i) $S$ denotes a finite set of states, (ii) $A(s)$ denotes
the finite set of actions available at state $s \in S$, (iii) $s_0$ is initial
state, (iv) $\envDist(s'\mid a, s)$ is the probability of transitioning from
$s$ to $s'$ when applying action $a\in A(s)$.

We will make two additional technical assumptions about $M$. First, we assume a
unique (always reachable) sink state, i.e., $\envDist(\eoe\mid a, \eoe) = 1$,
denoting ``end of episode''. Second, we shall assert the Luce choice axiom,
which requires that each action, $a \in A(s)$, be \emph{distinct}, i.e., no
actions are interchangeable or redundant at a given
state~\cite{luce1959individual}. Note that the Luce choice axiom is relaxed in
Sec~\ref{sec:luce_cegis}.

A \emph{path}, $\trc$, is an alternating sequence of states and
actions starting with $s_0$:
\begin{equation}
  \trc = s_0 \xrightarrow{a_0} s_1 \xrightarrow{a_1} s_2 \xrightarrow{a_2} \hdots.
\end{equation}
Any path, $\trc$, can be (non-uniquely) decomposed into a
\emph{prefix}, $\prefix$, concatenated with a \emph{suffix},
$\suffix$, denoted $\trc = \prefix\concat \suffix$. We allow $\suffix$
to be of length $0$. The last state of $\trc$ is denoted by
$\last(\trc)$. A path is \emph{complete} if it contains $\eoe$ exactly
once, and thus $\last(\trc) = \eoe$. We denote by
$\Path_\$$ the set of all complete paths, and by
$\Path$ the set of all prefixes of
$\Path_\$$, i.e., paths that contain $\eoe$ at most once.

\subsection{Task Specifications}
Next, we develop the machinery to describe the set of paths that
constitute performing a given task. In particular, a \emph{concept}
is a subset of paths,
\begin{equation}
  \concept \subseteq \Path_\$.
\end{equation}
We refer to a collection of concepts, $\ConceptClass$, as a
\emph{concept class}. 
Similarly a \textbf{representation class} is a set
$\RepresentationClass$ equipped with two maps:
\begin{equation}
    \text{encode} : \RepresentationClass \to \{0, 1\}^*\hspace{4em}
    \text{concept} : \RepresentationClass \to \ConceptClass,
\end{equation}
where $\text{encode}$ is injective and thus unambiguous.
The size or \textbf{description length} of a
representation is given by the length of its encoding:
\begin{equation}
    \size : \RepresentationClass \to \Nat \hspace{3em}
    \size(\spec) \eqdef |\encode(\spec)|.
\end{equation}
\begin{remark}
    While seemingly pedantic, the distinction between concept and
    representation proves immensely valuable when discussing the learnability
    of a concept and assigning prior probabilities on concepts, e.g., appealing
    to Occam's razor to order prior probabilities by size. Nevertheless, when
    clear from context, we shall often conflate representations and concept
    classes.
\end{remark}

A \emph{task specification} is thus formalized as a member of a
fixed representation class $\RepresentationClass$ with satisifaction
corresponding to set membership. Finally, to avoid notational clutter, when
clear from context, we shall often abuse notation and denote by $\spec$ a task
specification and its underlying concept, e.g., writing $\trc \in \spec$ rather
than $\trc \in \text{concept}(\spec)$.

\begin{example}\label{ex:conceptclass}
    We formalize the representation class of our motivating example.  Let (i)
    $\Sigma = \{\redTile, \blueTile, \yellowTile, \brownTile, \square\}$ denote
    an alphabet, (ii) $\spec_1$ and $\spec_2$ denote the left and right DFA
    over $\Sigma$ shown in Fig~\ref{fig:target-dfa}.  Define
    $\RepresentationClass_{\text{reg}}$ as the set of all tasks represented by a
    DFA over $\Sigma$, i.e. regular languages over $\Sigma$. We choose to
    encode DFA using a binary encoding of $\spec$'s multi-graph, i.e., encoding
    (i) the number of states $n$, (ii) the set of accepting (or rejecting)
    states, and (iii) the non-stuttering (not self loop) transitions, $m$. Note
    this encoding is dominated by (iii) with $\size(\spec) \in O(m \log n)$.
\end{example}

A \emph{labeled example} is a tuple, $\ex = (\trc, l)$, corresponding to
a complete path and a binary label, $l \in \{0, 1\}$. A collection of
labeled examples, $\EX = \ex_1, \ldots, \ex_n$, is \emph{consistent}
with a task, $\spec$, if:
\begin{equation}
  \forall x_i=(\trc_i, l_i)~.~(\trc_i \in \spec) \iff (l = 1).
\end{equation}

An \emph{identification algorithm}, $\identify$, maps a collection of
labeled examples, $\EX$ to a distribution over consistent tasks in
$\RepresentationClass$ or $\bot$ if no consistent task exists.
\begin{example}
  Let $\trc_b$ and $\trc_r$ be the completed black and red paths shown
  in Fig~\ref{fig:gw8x8} and define
  $\EX_{bg} = \{(\trc_b, 1), (\trc_r, 0)\}$. $\spec_2$ is consistent
  with $\EX_{bg}$ and $\spec_1$ is not.
\end{example}

\subsection{Policies and Demonstrations} A (history dependent)
\emph{policy}, $\policy(a\mid \trc)$, is a distribution over actions,
$a$, given a path, $\trc$, where $a \in A(\last(\trc))$.  A policy is
$(p, \spec)$-\emph{competent} if:
\begin{equation}
  \psat_\spec(\policy) \eqdef \Pr(\trc \in \spec \mid \policy, M) = p 
\end{equation}

A \emph{demonstration} is a path, $\demo$, generated by a employing a
policy $\policy$ in an MDP $M$, $\trc \sim (\policy, M)$.
\begin{mdframed}[backgroundcolor=blue!5,nobreak=true]
  \textbf{Task Inference from Demonstrations Problem} (TIDP): Let a
  $M$, $\RepresentationClass$, and $P$ be a fixed MDP, representation class, and
  task prior, respectively.  Further, let $\policy^*$ be a
  $(p^*, \spec^*)$-competent policy, $\policy^*$, where
  $p^*, \spec^*,$ and $\policy^*$ are all unknown.  Given a multi-set
  of i.i.d. demonstrations, $\Demos \sim (\policy^*, M)$, find:
  \begin{equation}
    \spec \in \argmax_{\psi \in \RepresentationClass} \Pr(\Demos \mid \psi, M) \cdot  P(\psi \mid M).
  \end{equation}
\end{mdframed}
Of course, by itself, the above formulation is ill-posed as
$\Pr(\Demos~|~M, \varphi)$ is left undefined. What remains is to
derive a suitable agent model and discuss how to manipulate
likelihoods in this model.

\section{Task Motivated Agents}\label{sec:agent_model}
Following~\cite{DBLP:conf/cav/Vazquez-Chanlatte20}, we
propose using the principle of maximum causal entropy
to assign a bias-minimizing belief of generating the demonstrations given a candidate task.

\subsection{Maximum Causal Entropy Policies} We start by defining the
causal entropy on arbitrary sequences of random variables.  Let
$\RV{X}_{1:i} \eqdef \RV{X}_1, \hdots, \RV{X}_i$ and
$\RV{Y}_{1:i} \eqdef \RV{Y}_1,\hdots,\RV{Y}_i$ denote two sequences of
random variables.  The \emph{entropy} of $\RV{X}_{1:i}$ \emph{causally
  conditioned} on $\RV{Y}_{1:i}$ is:
\begin{equation}
  H(\RV{X}_{1:i}\mid\mid \RV{Y}_{1:i}) \eqdef \sum_t^i H(\RV{X}_{t}~|~\RV{Y}_{1:t}, \RV{X}_{1:{t-1}})
\end{equation}
where,
$H(\RV{X} \mid \RV{Y}) \eqdef \E_{\RV{X}}[-\ln\Pr(\RV{X}~|~\RV{Y})]$,
denotes the entropy of $\RV{X}$ (statically) conditioned on $\RV{Y}$.
Intuitively, causal conditioning enforces that past variables do not
condition on events in the future. This makes causal entropy
particularly well suited for robust forecasting in \emph{sequential}
decision making problems, as agents typically cannot observe the
future~\cite{mce}.

For MDPs, the unique policy, $\policy_\spec$, that maximizes entropy
subject (i) a finite horizon and (ii) to being $(p, \spec)$-competent
exponentially biases towards higher value actions~\cite{mce}:
\begin{equation}\label{eq:softbackup}
  \ln \pi_\lambda(a\mid \trc) \eqdef Q_\lambda(\trc\concat a) - V_\lambda(\trc)
\end{equation}
where the values are recursively given by the following smoothed
Bellman-backup:
\begin{equation}\label{eq:softbackup2}
  Q_\lambda(\trc\concat a) \eqdef \E_{s}\big[V_\lambda(\trc\concat a\concat s) \mid a, \trc\big],
\end{equation}
\begin{equation}
  V_\lambda(\trc) \eqdef
  \begin{cases}
    \lambda \cdot [\trc \in \spec] & \text{if } \trc \text{ is complete}\\
    \displaystyle \LSE_{a \in A(\last(\trc))} Q_\lambda(\trc\concat a) &
    \text{otherwise}
  \end{cases}.
\end{equation}
Here $\LSE_x f(x) \eqdef \ln\sum_xe^{f(x)} $ and $\lambda$, called the
\emph{rationality}, is set such that
$\psat_\spec(\policy_\lambda) = p$. When $\lambda$ is induced from
$\spec$, we will often write $V_\spec$, and $Q_\spec$.

\begin{remark}
  $\psat_\spec(\policy_\lambda)$ increases monotonically in $\lambda$
  and thus can be efficiently calculated to arbitrary precision using
  binary search.
\end{remark}

\begin{remark}
  The competency of the agent can be treated as a hyper-parameter or
  estimated empirically, e.g.,
  $p_\spec \approx \nicefrac{1}{m}\sum_{i=1}^m [\trc \in \spec]$. The
  former is useful when given on a few demonstrations and the latter
  is useful when given a large number of demonstrations.
\end{remark}

\subsection{Explainability of a task}
Next, we develop machinery to measure how well a task explains the
demonstrators behavior. To this end, the \emph{surprise}\footnote{also known as
surprisal or information content} of a demonstration is defined as:
\begin{equation}
  \surprise(\trc \mid \policy, M) \eqdef -\ln \Pr(\trc\mid \policy, M).
\end{equation}
Intuitively, the surpise correlates with the number of bits (or nats) required
to encode $\trc$ given minimizing expected description lengths under $(\policy,
M)$ ~\cite{cover2012elements}.

The surprise of a collection of demonstrations is the sum of the
surprise of each demonstration:
\begin{equation}
  \surprise(\Demos \mid \policy, M) \eqdef
  \sum_{i=1}^m\surprise(\demo_i\mid \policy, M)  .
\end{equation}
Note that the likelihood of i.i.d. demonstrations from $(\policy, M)$
is simply,
\begin{equation}
    \exp(-\surprise(\Demos )).
\end{equation}
Given a \emph{fixed} MDP, $M$,
and a \emph{fixed} collection of demonstrations,
$\trc_1, \ldots, \trc_m$, we define the surprise of a task, $\spec$,
as:
\begin{equation}
  \surprise(\spec) \eqdef \surprise(\Demos \mid \policy_\spec, M) 
\end{equation}
Thus, solving a TIDP requires minimizing $h$ and the negative log
prior, which w.l.o.g. can be taken as $\text{size}(\spec)$.

\section{Manipulating Surprise}\label{sec:sggs}
In the sequel, we seek to study how changing a task $\spec$ changes
the corresponding surprise, $\surprise(\spec)$, and thus the
likelihood of observing the demonstrations.

\subsection{Prefix Tree Perspective} 
\begin{figure}[h]
  \centering \scalebox{0.65}{ 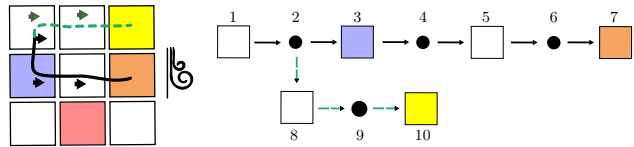 }

  \caption{Prefix tree with 10 nodes for the paths shown on the left.}
  \label{fig:prefix_tree}
\end{figure}
To facilitate this, we will find
it useful to re-frame paths as either \emph{deviating} or
\emph{conforming} to the demonstrations. To start, denote by
$\prefixtree = (\Nodes, \Edges)$ the \emph{prefix tree} (or trie) of
the demonstrations, $\Demos$, where $\Nodes$ and $\Edges$ are the
\emph{nodes} and \emph{edges} of $\prefixtree$, respectively. Each
node $i \in \Nodes$, corresponds to a prefix, $\prefix_i$, of at least
one of the demonstrations.  An edge connects \emph{parent} $i$ to
child $c$ if $\prefix_c$ is the one action (or state) extension of
$\prefix_i$, i.e., $\prefix_c = \prefix_i \sigma$, where $\sigma$ is a
path of length 1.  For each edge, $(i, j) \in \Edges$, we define the
\emph{edge traversal count}, $\#_{(i, j)}$, as the number of times
prefix $\prefix_j$ appears in the demonstrations. The set of unique
demonstrations maps directly to the leaves of $\prefixtree$, i.e., for
each leaf $v$, $\prefix_v$ is a demonstration. A node is said to be an
\emph{ego node} if it corresponds to selecting an action, and an
\emph{environment (env) node} otherwise.

We say a path, $\trc$, \emph{conforms} to the demonstrations if there
is a node $i$ such that, $\prefix_i = \trc$. A path \emph{deviates}
from the demonstrations if it is not conforming. The \emph{pivot} of a
deviating path, $\trc$, is the node corresponding to the longest
prefix of $\trc$ that conforms to the demonstrations. Note that it is
possible to pivot at the leaves of the tree, i.e., the longest prefix
is a demonstration.

\begin{example}
  Example demonstrations and the corresponding prefix tree are
  illustrated in Fig~\ref{fig:prefix_tree}. Note that it is possible
  to pivot at every node \emph{except} node 2, since both
  possibilities (slipping/not slipping) appear in the demonstrations.
\end{example}

\subsection{Change of variables} 
\begin{figure}
  \centering \scalebox{0.9}{ 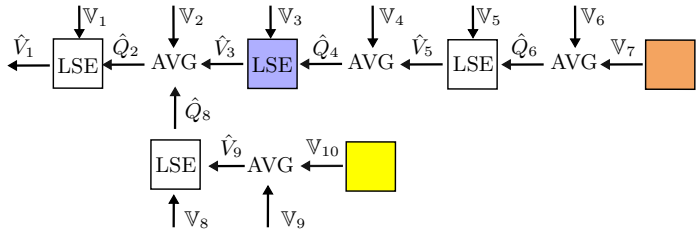
  }

  \caption{Computation tree of $\hat{Q}$ and $\hat{V}$ values for each
    node of prefix tree given by soft Bellman
    backup~\eqref{softbackup2} and pivot values, $\pival$.}
  \label{fig:computation_tree}
\end{figure}

Next, observe that because weighted
averaging and LSE are commutative, one can aggregate the values of a
set of actions or set of states (environment actions). In particular,
let $A_{i}$ and $S_{j}$ denote the \emph{conforming actions} and
\emph{conforming states} at an ego node $i$ and an env node $j$
respectively. The \emph{pivot value} of a node $i$ is:
\begin{equation}
  \pival_{i}^\spec \eqdef
  \begin{cases}
    \LSE_{a \notin A_{i}} Q_\spec(\prefix_i\concat a) & \text{if } i \text{ is ego},\\
    \E_s[V_\spec(\prefix_i\concat s) ~|~s \notin S_i, \prefix_i,  M] & \text{if } i \text{ is env},\\
  \end{cases}
\end{equation}
We shall denote by $\pival^\spec \in \Reals^{\Nodes}$ the node-indexed
vector of pivot values associated with task $\spec$ under our maximum
entropy agent model.
Crucially, the pivot values {\bf entirely
  determine} (see Fig~\ref{fig:computation_tree}) the values of the
states and actions visited by the demonstrations
via~\eqref{softbackup}.
%
Namely, let $\hat{Q}_k(\pival)$ and $\hat{V}_k(\pival)$ denote the
derived action and state value at node $k$ in the prefix tree, and let
$\Pr(i \rightsquigarrow k\mid \pival)$ denote the probability of
transitioning from node $i$ to node $k$ under the (local) policy:
$e^{\hat{Q}_j(\pival) - \hat{V}_i(\pival)}$. We refer to the surprise
map over pivot values as the \emph{proxy surprise}, $\hat{\surprise}$,
i.e.:
\begin{equation}\label{eq:proxy}
  \hat{\surprise}(\pival) \eqdef -\hspace{-1em}\sum_{(i, j) \in \Edges}\hspace{-0.6em}\#_{(i, j)} \cdot \ln \Pr(i \rightsquigarrow j\mid \pival).
\end{equation}

\subsection{Leveraging proxy gradients} Unlike the surprise,
$\surprise(\spec)$, the \emph{proxy surprise}, $\hat{\surprise}$, is
differentiable with respect to the pivot values $\pival$. One can
interpret $\nabla \hat{\surprise}(\pival^\spec)$ then as suggesting
how to modify the pivot values in order to make the demonstrations
less surprising. A natural question then is how to adapt $\spec$ given
this knowledge. Observe the following two propositions (with proof
sketches of the mechanical calculations provided in Sec~\ref{sec:proofs}):
\begin{proposition}[Pivot values respect subsets]\label{prop:monotone}
  Let $\trc$ be a complete path that pivots at node $i$. If
  $\spec \subsetneq \psi$ and $\trc \in \psi \setminus \spec$, then
  $\pival^\spec_i < \pival^\psi_i$.
\end{proposition}
\begin{proposition}[$\nabla \hat{\surprise}$ determined by path
  probabilities]\label{prop:grad}
  Let $p_{xy}(\pival)$ denote the probability of starting at node $x$ and pivoting at $y$, i.e.,
 \begin{equation}
     p_{xy}(\pival) = \Pr(x \rightsquigarrow y)\cdot \bigg(1 - \sum_{\substack{z\\(y, z) \in E }} \Pr(y \rightsquigarrow z)\bigg)
 \end{equation} 
  then,
  
  \begin{equation}\label{eq:grad}
    \small
    \frac{\partial \hat{\surprise}}{\partial \pival_k} =
    \hspace{-0.5em}
    \sum_{\substack{(i, j) \in \Edges\\i \text{ is ego}}}  \hspace{-0.5em}\#_{(i, j)} \cdot\bigg( p_{ik}(\pival) - p_{jk}(\pival)\bigg)
  \end{equation}
\end{proposition}
Prop~\ref{prop:monotone} suggests that to adjust the pivot value,
$\pival_i$, in a manner that decreases surprise, one can simply take a
path, $\trc$, that pivots through $i$ such that:
\begin{equation}\label{eq:toggle}
  \trc \in \spec \iff \frac{\partial \hat{\surprise}}{\partial \pival_i} > 0,
\end{equation}
and negate its satisfaction under $\spec$.

Prop~\ref{prop:grad} illustrates that (i) the gradients are simple to
compute given access to the policy on the prefix tree and (ii)
sampling from $\policy_\spec$ yields paths with a large effect on the
gradient. To see point (ii), consider extending the prefix tree to
contain the most likely paths after pivoting and apply
Prop~\ref{prop:grad}. The gradient w.r.t. the pivot value
corresponding to the newly introduced leaves is their path probability
and thus their probability of being sampled after pivoting!

\subsubsection{Relation to heuristics}
Eq~\eqref{grad} captures several intuitive heuristics for changing
$\pival$ to make the demonstrations more likely.
Consider ego edge $(i, j)$, which corresponds to
an action the agent took. Pivoting at $i$, i.e., $k=i$ above, yields,
$p_{jk}(\pival)= 0$. Thus, edge $(i, j)$ contributes positively to the gradient.
Since we want to minimize surprisal, this suggests that we
want a to decrease $\pival_k$ and thus make the action the agent took look
more valuable by comparison. Similarly, suppose $k = j$, implying that: \[
    p_{ik} - p_{jk} = \Pr(i \rightsquigarrow j\mid \pival) \cdot p_{jk} - p_{jk} \leq 0, \]
where equality holds iff exactly one action is available.
Thus, when $k=j$, edge $(i, j)$ typically provides a negative contribution to the surprisal gradient.
Again, because we want to minimize surprisal, the above suggests an increase in $\pival_k$,
and by extension, an increase in the value of the action used on $(i, j)$.
These two cases are summarized by the following rules of thumb:
\begin{mdframed}
    \begin{enumerate}
        \item Make the actions taken more optimal by decreasing the value of other actions.
        \item Make the actions taken less risky by increasing the value of possible outcomes.
    \end{enumerate}
\end{mdframed}

\subsubsection{Surprise Guided Sampling}
\begin{figure}[t]
  \centering \scalebox{0.8}{ 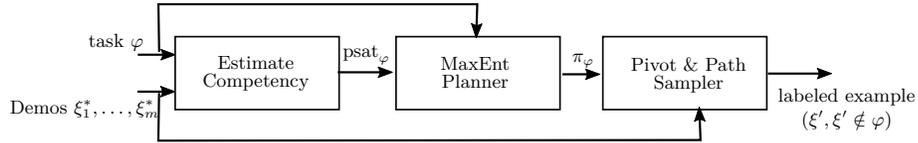 }

  \caption{Overview of surprise guided sampler.}
  \label{fig:sgs}
\end{figure}

Using these insights we propose surprise guided sampling
(Alg~\ref{alg:sgs} and Fig~\ref{fig:sgs}) which samples a path to
relabel based on (i) how likely it is under $\pi_\spec$ and (ii) the
magnitude and sign of the gradient at the corresponding pivot.
Combined with an identification algorithm, $\identify$, repeated
applications of Alg~\ref{alg:sgs} yields an infinite (and stochastic)
sequence of tasks resulting from incrementally conjecturing
mis-labeled paths.

\begin{algorithm}[H]
  \caption{Surprise Guided Sampler (SGS)\label{alg:sgs}}
  \begin{algorithmic}[1]
    \State {\bfseries Input:} {$\spec, \EX, \prefixtree, M, \beta$}
    \State Compute $\policy_{\spec}$ given $M$ and $\prefixtree$.
%
    \State Let $D$ be the distribution over pivots s.t.
    \[\Pr(\text{pivot }i) \propto \exp{\left(-\frac{1}{\beta}\left |
          \frac{\partial \hat{\surprise}}{\partial \pival_i}\right|
      \right)}.\]
    \State Sample a pivot $i \sim D$ and a path
    $\trc \sim (\policy_\spec, M)$ s.t.
    \begin{enumerate}[noitemsep,label=\roman*]
    \item $\trc$ pivots at $i$.
    \item
      $\trc \in \spec \iff\frac{\partial \hat{\surprise}}{\partial
        \pival_i} > 0$.
    \item $\exists \spec' \in \RepresentationClass$ s.t. $\spec'$ is
      consistent with:
      \[\hspace{-5.5em}\EX \cup \{(\trc, \trc \notin \spec)\}.\]
    \end{enumerate}
    \State \textbf{return} $\trc$
  \end{algorithmic}
\end{algorithm}

\begin{remark}
  Alg~\ref{alg:sgs} only \emph{requires} a black box maximum entropy
  (MaxEnt) planner to enable assigning edge probabilities,
  $\Pr(i \rightsquigarrow j\mid \pival)$, and sampling suffixes given
  a pivot. If the satisfaction probability of an action is also known,
  i.e.,
  $\Pr_{\xi'}(\trc\concat \trc' \in \spec \mid\trc, M, \policy_\spec)$,
  then one can more efficiently sample suffixes using Baye's rule and
  the policy $\policy_\spec$.
\end{remark}

\begin{example}\label{ex:sgs}
  Recall our ad-hoc analysis on Fig~\ref{fig:gw8x8} in
  Sec~\ref{robottaskspec}. Under the reach $\yellowTile$ while
  avoiding $\redTile$ hypothesis, $\spec$, it is surprising that the
  agent moves up from $\blueTile$ rather than following the red dashed
  suffix. That is, there is a non-trivial probability to
  \textit{deviate} at this point, call $k$. Using a planning horizon
  of 15 steps, $\frac{\partial \hat{\surprise}}{\partial \pival_k}$ is
  positive and indeed larger in magnitude than all other pivots.
  Props~\ref{prop:monotone} and~\ref{prop:grad} suggest sampling
  (using $\policy_\spec$) a new path, $\trc$, that pivots from the
  demonstration at $k$ and satisfies $\spec$.  Note that the
  illustrated red dashed suffix indeed fits this description.
  Finally, Eq~\ref{eq:toggle} prescribes marking $\trc$ as a negative
  path which matches our ad-hoc analysis!
\end{example}

%
%

\section{Specification Search}\label{sec:DISS}

In this section, we take the insights developed in the previous
sections and propose the DISS algorithm - a variant of simulated
annealing for quickly finding probable task specifications.
\subsection{Intuition}
DISS is a random walk through labeled example space guided by the
surprise gradient of sampled concepts. Each new labeled example helps
constrain the concept sampler to make previously ``surprising'' paths
less surprising. As we shall see, this accumulation of corrective
constraints systematically focuses the concept sampler on more and
more probable (lower energy) task specifications. Simulated Annealing
formalizes this process, with periodic resets helping to avoid
over-committing to a particular labeled example.

\subsection{Background on Simulated Annealing}
At a high level, \emph{Simulated Annealing}
(SA)~\cite{DBLP:conf/wsc/SkiscimG83} is a probabilistic optimization
method that seeks to minimize an energy function
$U : Z \to \Reals \cup \{\infty\}$. To run SA, one requires three
ingredients: (i) a \emph{cooling schedule} which determines a
monotonically decreasing sequence of temperatures, (ii) a
\emph{proposal} (neighbor) distribution $q(z'\mid z)$, and (iii) a
\emph{reset} schedule, which periodically sets the current state,
$z_t$, to one of the lowest energy candidates seen so far.

A standard simulated annealing algorithm then operates as follows: (i)
An initial $z_0 \in Z$ is selected (ii) $\temperature_t$ is selected
based on the cooling schedule. (iii) A neighbor $z'$ is sampled from
$q(\bullet\mid z_t)$. (iv) $z'$ is accepted ($z_{t+1} \gets z'$) with
probability:
\begin{equation}
  \small
  \Pr(\text{accept}\mid z', z_t) = \begin{cases}
    1 & \text{if } dU > 0\\
    \min\left\{1, e^{\nicefrac{dU}{\temperature_t}}\right\} & \text{otherwise}\\
  \end{cases},
\end{equation}
where $dU \eqdef U(z) - U(z')$. (v) Finally, if a reset set is
trigger, $z_{t+1}$ is sampled from previous candidates, e.g., uniform
on the argmin.

As previously stated, we propose a variant of simulated annealing
adapted for our specification inference problem.  We will start by
assuming the posterior distribution on tasks takes the form:
\begin{equation}
  \Pr(\spec\mid \Demos, M) \propto e^{-U(\spec)},
\end{equation}
where the \emph{energy}, $U$, is given by:
\begin{equation}
  U(\spec) \eqdef \size(\spec) + \theta\cdot \surprise(\spec),
\end{equation}
where $\theta \in \Reals$ determines the relative weight of the
surprise.  That is, we appeal to Occam's razor and assert that the
task distribution is exponentially biased towards simpler tasks, where
simplicity is measured by the description length of the task,
$\size(\spec)$, and the description length (i.e. surprise) of $\Demos$
under $(\policy_\spec, M)$.

\subsection{Algorithm Description}
Using the language of SA, we define DISS as follows: (i) $z \in Z$ is
a tuple, $(\EX, \spec)$, of labeled examples and a task
specification. (ii) $z_0 = (\emptyset, \bot)$. (iii) the proposal
distribution, $q(\EX', \spec'\mid\EX, \spec)$ is defined to first
sample a concept using an identification map,
$\spec' \sim \identify(\EX)$, then run SGS on $\spec'$ to conjecture a
labeled path $\trc$, yielding
$\EX' = \EX \cup \{(\trc, \trc \notin \spec')\}$. (iv) resets occur
every $\kappa \in \Nat$ time steps. If a reset is triggered,
$\EX_{t+1}$ is sampled from $\text{softmin}_{i \leq t} U(\spec_i)$,
and $\spec_{t+1}$ is sampled from $\identify(\EX_{t+1})$.  This
algorthm is given as psuedo code in Appendix~\ref{sec:diss-pseudo}.

\section{Experiments}\label{sec:experiment}


In this section, we illustrate the effectiveness of DISS by having it
search for a ground truth specification, represented as a DFA, given
the expert demonstrations, $\trc_b, \trc_g$, from our motivating
example (shown in Fig~\ref{fig:gw8x8}).  The (dotted) green path,
$\trc_g$, goes directly $\yellowTile$. The (solid) black path,
$\trc_b$, immediately slips into $\blueTile$, visits $\brownTile$,
then proceeds towards $\yellowTile$.  This path is incomplete, with a
possible extension, $\suffix_b$, shown as a dotted line. The ground
truth task is the right DFA in Fig~\ref{fig:target-dfa}.


We consider two specification inference problems by varying the
representation class and the provided demonstrations. These variants
respectively illustrate that (i) our method can be used to
incrementally learn specifications from unlabeled incomplete
demonstrations and (ii) the full specification can be learned given
unlabeled complete demonstrations.
\begin{enumerate}[noitemsep]
\item \textbf{Monolithic}: $\trc_g$ and $\trc_b\concat\suffix_b$ are provided
    as (unlabeled) \emph{complete} demonstrations. The representation class is
    the set of DFAs over $\{\redTile, \yellowTile, \blueTile,
    \brownTile\}$, e.g., $\RepresentationClass_{reg}$ from
    Example~\ref{ex:conceptclass}.

\item \textbf{Incremental}: $\trc_b$ is provided as an (unlabeled)
    \emph{incomplete} demonstration. The representation class, $\RepresentationClass \subsetneq \RepresentationClass_{reg}$ is the set of DFAs that require recharging and avoiding lava.

\end{enumerate}
The surprise weight, $\theta$, is set to $1$ for both variants.
Finally, two additional inductive biases, which empirically proved
necessary for optimizing the random pivot baseline, are applied: (i)
we remove white tiles, $\square$, from labeled examples (ii) we
transform sequences of repeated colors into a single color thus
biasing towards DFA that do not count. For example,
$\square\blueTile\blueTile\square\yellowTile\blueTile \mapsto
\blueTile\yellowTile\blueTile$.

\begin{figure}[t]

\begin{subfigure}[t]{0.48\textwidth}
  \centering \scalebox{0.43}{
    \import{imgs/}{mass_mono.pgf} }
  \caption{Min energy found by iteration in monolithic experiment.}
  \label{fig:mass_mono}
\end{subfigure}
\hfill
\begin{subfigure}[t]{0.48\textwidth}
  \centering \scalebox{0.43}{
    \import{imgs/}{mass_inc.pgf} }
  \caption{Min energy found by iteration in incremental experiment.}
  \label{fig:mass_inc}
\end{subfigure}
\end{figure}

\textbf{DISS parameters.} Our implementation of DISS uses SGS
temperature $\beta = \nicefrac{1}{500}$, resets every $10$ iterations,
and uses the following cooling schedule:
\begin{equation}
  \temperature_t = 100\cdot(1 - \nicefrac{t}{100}) + 1.
\end{equation}
For concept identification, $\identify$, we adapt an existing
SAT-based DFA identification
algorithm~\cite{DBLP:conf/lata/UlyantsevZS15} to enumerate the first
20 consistent DFAs (ordered by size).  A DFA is then sampled from
$\text{softmin}_\spec(\text{size}(\spec))$. For maximum entropy
planning, we use the Binary Decision Diagram based approach proposed
in~\cite{DBLP:conf/cav/Vazquez-Chanlatte20} with a planning horizon of
15 steps.  Finally, because our experiments operate with one or two
demonstrations, the rationality coefficent, $\lambda$, was fixed to 10. 
Nearly identical results we also achieved by setting
$\lambda$ such that the competency, $p_\spec$ is as close to 
$\nicefrac{4}{5}$ as possible. 

\textbf{Baselines.}  As mentioned in the introduction, existing techniques for
learning specifications from demonstrations use various \emph{syntactic}
concept classes, each with their own inductive
biases~\cite{%
    DBLP:conf/nips/Vazquez-Chanlatte18,%
    DBLP:conf/nips/ShahKSL18,%
    DBLP:conf/icra/Yoon021%
}. Thus, we implemented two DFA adapted baselines that act as proxies for the
enumerative and probabilistic hill climbing style algorithms of existing work:

\begin{enumerate}[noitemsep]
\item \textbf{Prior-ordered Enumeration.} This baseline uses the same
  SAT-based DFA identification algorithm to find the first $N$ DFAs,
  ordered by prior probability, i.e., size.  As an alternative to
  DISS's competency assumption, we allow the enumerative baseline to
  restrict the search to task specifications that accept the provided
  demonstrations.\footnote{For the incremental experiment, a
    counterexample loop is used to add labeled examples that bias the
    DFAs to imply $\spec_1$.}

\item \textbf{Random Pivot DISS.}  This baseline uses DISS with SGS
  temperature, $\beta=\infty$. This ablation results in a (labeled
  example) mutation based search with access to the same class of
  mutations as DISS, but samples pivots uniformly at random, i.e., no
  gradient based bias.  Note that this variant still samples suffixes
  conditioned on the sign of the gradient, and thus the mutations are
  still informed by the surprise.
\end{enumerate}

\subsection{Results and Analysis}

To simplify our analysis, we present time in iterations, i.e., number
of sampled DFAs, rather than wall clock time. This is for two
reasons. First, for each algorithm, the wall clock-time was dominated
by synthesizing maximum entropy planners for each unique DFA
discovered, but the choice of planner is ultimately an implementation
detail. Second, because many DISS iterations correspond to the same
DFAs (due to resets and rejections) the enumeration baseline explored
significantly more \emph{unique} DFAs than DISS (a similar effect
occurs with the random pivot baseline, since the different pivots give
more diverse example sets). Thus, using wall clock-time would skew the
results below in DISS's favor.\footnote{ Nevertheless, for the
  monolothic experiment, the wall clock times for DISS, random
  pivoting, and enumeration were 542s, 764s, and 617s respectively.
  Similarly, for the incremental experiment the respective times were
  353s, 464s, and 820s.  }

Fig~\ref{fig:mass_mono} and Fig~\ref{fig:mass_inc} show the minimum
energy DFA for the monolithic and incremental experiments
respectively.  We see that for both experiments, DISS was able to
significantly outperform the baselines (recall that energy is the
negative log of the probability), with the incremental experiment
requiring only a few (< 6) iterations to find a probable DFA!
Furthermore, in addition to finding the most probable DFAs much faster
than the baselines, DISS also found \emph{more} high probability DFAs.

\begin{figure}
  \centering \scalebox{0.38}{ 
\begingroup%
  \makeatletter%
  \providecommand\color[2][]{%
    \errmessage{(Inkscape) Color is used for the text in Inkscape, but the package 'color.sty' is not loaded}%
    \renewcommand\color[2][]{}%
  }%
  \providecommand\transparent[1]{%
    \errmessage{(Inkscape) Transparency is used (non-zero) for the text in Inkscape, but the package 'transparent.sty' is not loaded}%
    \renewcommand\transparent[1]{}%
  }%
  \providecommand\rotatebox[2]{#2}%
  \newcommand*\fsize{\dimexpr\f@size pt\relax}%
  \newcommand*\lineheight[1]{\fontsize{\fsize}{#1\fsize}\selectfont}%
  \ifx\svgwidth\undefined%
    \setlength{\unitlength}{586.57463494bp}%
    \ifx\svgscale\undefined%
      \relax%
    \else%
      \setlength{\unitlength}{\unitlength * \real{\svgscale}}%
    \fi%
  \else%
    \setlength{\unitlength}{\svgwidth}%
  \fi%
  \global\let\svgwidth\undefined%
  \global\let\svgscale\undefined%
  \makeatother%
  \begin{picture}(1,0.38873547)%
    \lineheight{1}%
    \setlength\tabcolsep{0pt}%
    \put(0,0){\includegraphics[width=\unitlength,page=1]{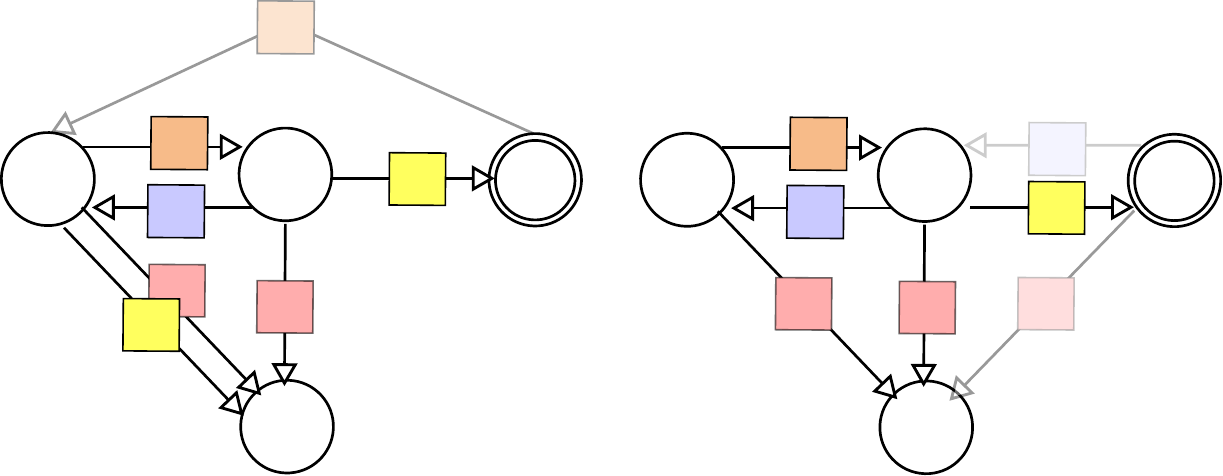}}%
    \put(0.75639619,0.23996418){\makebox(0,0)[t]{\lineheight{1.25}\smash{\begin{tabular}[t]{c}start\end{tabular}}}}%
    \put(0.23508091,0.24071011){\makebox(0,0)[t]{\lineheight{1.25}\smash{\begin{tabular}[t]{c}start\end{tabular}}}}%
  \end{picture}%
\endgroup%
 }
  \caption{Most probable DFA found by DISS in the monolithic
    experiment (left) and the incremental experiment (right).}
  \label{fig:learned_dfas}
\end{figure}

Next, we analyze the DFAs learned by DISS.  The most likely DFAs found
by DISS for each experiment (left and right) are shown in
Fig~\ref{fig:learned_dfas}.  We observe that for both experiments,
DISS is able to learn that if the agent visits $\blueTile$, it needs
to visit $\brownTile$ before $\yellowTile$!  Furthermore, the
monolithic DFA is impressively able to learn that $\redTile$ leads to
a sink state, a feat that requires quite a number of negative examples
to illicit from our size-based DFA sampler.  In fact, this discovery
is responsible for the large drop in energy at 15 iterations in the
monolithic experiment.

Nevertheless, our learned DFAs differ from ground truth, particularly
when it comes to the acceptance of strings \emph{after} visiting
$\yellowTile$. We note that a large reason for this is that our domain
and planning horizon make the left most $\yellowTile$ effectively act
as a sink state. That is, the resulting sequences are
indistinguishable, with many even having the exact same energy.  In
Fig~\ref{fig:learned_dfas}, we make such edges lighter, and note that
the remainder of the DFAs show good agreement with the ground
truth. Finally, while impressive, this points to a fundamental
limitation of demonstrations. Namely, if two tasks have very
correlated policies in a workspace, then without strong priors or side
information, one is unable to distinguish the tasks. For example, if a
task requires the agent to avoid $\redTile$, but no $\redTile$ are
shown in the workspace, then one cannot hope to learn this aspect of
the task.

\section{Relaxing the Luce axiom}\label{sec:luce_cegis}
Our design and analysis of DISS assumed the Luce axiom and in particular that
all actions accessible by states in the demonstrations are distinct, i.e.,
if for all non-terminal states $s$,
\[
    \forall a_1, a_2 \in A(s)~.~ a_1 \neq a_2 \implies \delta(s, a_1) \neq \delta(s, a_2).
\]
This assumption is important several reasons. Chiefly, a violation
of the Luce axiom along the demonstrations means that the pivot values
are no longer independent. We illustrate with an example.
\begin{example}
    Consider the demonstration $\trc = s_0\xrightarrow{a_2} s_1$, where $A(s_0)
    = \{a_1, a_2\}$ and $A(s_1) \neq \emptyset$. Thus, for prefix $\prefix = s_0$,
    we have $a_1 \in A_\prefix$ and $a_2 \notin A_\prefix$. Since $a_1$ is the
    only action that pivots, we have $\pival_{\prefix} = V(\prefix\concat a_1)$. 
    Similarly, since $A(s_1) \neq \emptyset$, $\trc$ is not terminal and has
    its pivot value determined be the subtree accessed by $\prefix \concat a_2$, i.e.,
    $\pival_\trc = \pival_{\prefix \concat a_2} = V(\prefix \concat a_2)$.
    But since $a_2$ accesses the same subtree as $a_1$, then \[
        \pival_\prefix = V(\prefix \concat a_1) = V(\prefix \concat a_2) = \pival_{\prefix \concat a_2} .\]
    Thus, $\pival_{\prefix} = \pival_{\prefix\concat a_2}$, implying the that pivot values are not
    independent.
\end{example}

\noindent
This has serious implications on our use of the surprise gradient.
The most glaring issue is that two logically equivalent pivots
may have opposing signs in the gradient. This can lead to conjecturing a
counter-factual that is exactly counter to your goal. For example,
if the last action in a positive demonstration is redundant with another action,
pivoting at that point will likely result in the demonstration being marked
negative! While it is possible for DISS to correct this conjecture in a future iteration,
this can seriously degrade performance in practice. Furthermore, 
a similar argument can be made for redundant states!

\subsection{Counter Example Guided Action Refinement}
For these reasons, it behooves us to develop a way to handle redundant actions
and states. Luckily there is a simple solution that only requires a minor
modification to DISS:
\begin{enumerate}
    \item Start by assuming that all (time-indexed) states and actions in the demonstrations are
          equivalent. Thus, the prefix tree is initially a chain and no nodes can be pivots.
      \item Whenever proof is found that two states (actions) are distinct at a given
          node in the prefix tree, create a new prefix tree where the prefix indexed equivalence
          class of states (actions) is partitioned accordingly.
\end{enumerate}
To formalize this algorithm, we introduce the idea of value-distinguishability.
\begin{mdframed}
    Let $\trc$ and $\trc'$ be two paths such that $\trc = \prefix\concat x\concat y$ and $\trc' = \prefix \concat x'\concat y'$,
    where $|x| = |x'| = 1$. Given a subset of the representation class $\Psi \subseteq \RepresentationClass$,
    we say that prefixes $\prefix\concat x$ and $\prefix\concat x'$ are $\Psi$-\textbf{value-distinguishable} 
    if there exists a task, $\spec\in\Psi$, such that:
    \begin{equation}\label{eq:luce_effective}
        V_\spec(\prefix\concat x) \neq V_\spec(\prefix\concat x').
    \end{equation}
    When $\Psi = \RepresentationClass$, we suppress $\Psi$ and simply say value-distinguishable.
\end{mdframed}
Importantly, if two prefixes are value distinguishable, then they maintain the
independence of pivot values since they must not access the same subtree.
Conversely, if two prefixes are not value distinguishable, then they access
must access functionally equivalent sub-trees. Thus, value-distinguishability
fits the needs of proof that two states (actions) are distinct.

The problem of course is that refuting value-distinguishability requires
examining the whole representation class (which may be impossible). Instead,
we define a series of equivalence relations based on an expanding
sets of specifications, $\Psi_1 \subseteq \Psi_2 \subseteq \ldots $,
where $\Psi_i$ is the set of specifications visited by DISS by iteration
$i$. Thus, at iteration $i$, prefix $\prefix$ and $\prefix'$ are in the 
same equivalence class if they are not $\Psi_i$-value-distinguishable.

In summary, to account for the possibility of redundant states and actions,
maintain an series equivalence relations. Whenever
two actions are deemed to be value-distinguishable, the equivalence relation
is updated, which results in a new prefix tree over representatives of
each equivalence class. Noting that the initial set of visited specifications
is empty and that no prefixes are $\emptyset$-value-distinguishable, we have
that (i) initial prefix tree is a chain; and (ii) no nodes can (initially) be pivots.

\begin{remark}
    This variant of DISS requires knowing the set of states and actions
    reachable from each state. This can be further relaxed by dynamically
    testing states and actions for value-distinguishability
    as they observed, e.g., by sampling. While not maintaining pivot value
    independence, it does prevent conjecturing conflicting labels to equivalent
    paths.
\end{remark}

\section{Conclusion}

This paper considered the problem of learning history dependent task
specifications, e.g. automata and temporal logic, from expert
demonstrations. We empirically demonstrated how to efficiently
explore intractably large representation classes such as deterministic finite
automata for find probable task specifications. The proposed family of
algorithms, \emph{Demonstration Informed Specification Search (DISS)},
requires only \emph{black box} access to (i) a Maximum Entropy planner
and (ii) an algorithm for identifying concepts, e.g., automata, from
labeled examples. While we showed concrete examples for the efficacy
of this approach, several future research directions remain. First and
foremost, research into faster and model-free approximations of
maximum entropy planners would enable a much larger range of
applications and domains.  Similarly, while large, the demonstrated
representation class was over a small number of pre-defined atomic
predicates. Future work thus includes generalizing to large symbolic
alphabets and studying more expressive specification formalisms such
as register automata, push-down automata, and (synchronous) products
of automata.

%
%
%
%
\newpage
\bibliographystyle{splncs04}
\bibliography{refs}
\appendix
\section{Proof Sketches}\label{sec:proofs}
\begin{proof}[Prop~\ref{prop:monotone}]
  Follows inductively from the monotonicity of $\E$, $\sum$, and
  $\ln$.
\end{proof}

Before proving Prop~\ref{prop:grad} we first prove the following
lemma.
\begin{lemma}\label{lem:grad}
  For any node, $i$, in the prefix tree,
  \[\frac{\partial}{\partial \pival_k} \blacksquare_i(\pival) = p_{ik}(\pival),\] where $\blacksquare$ denotes
  $\hat{V}$ for ego nodes and $\hat{Q}$ for env nodes.
\end{lemma}
\begin{proof}
  To begin, note that one can extend the prefix tree with new leaves
  corresponding to the action of pivoting at the node.
  Thus, $p_{ik}$, is simply $\Pr(i \rightsquigarrow k')$,
  where $k'$ is the pivot leaf for $k$. Thus, it is sufficient to prove that in this new tree:
  
  \[\frac{\partial}{\partial \pival_k} \blacksquare_i(\pival) = \Pr(i \rightsquigarrow k \mid \pival),\]
  
  For any edge $(a, b)$, observe that if $a$ is an environment node,
  then $\Pr(a\rightsquigarrow b\mid \pival)$ is a constant, denoted
  $q_{ab}$. Next, observe that because the nodes are arranged as a
  tree either: (1) $k$ is not reachable from $i$ or (2) only a single
  edge, call $(i, j)$, can reach $k$ from $i$. Thus,
  \begin{equation}\label{eq:lem1}
    \begin{split}
      \frac{\partial \hat{Q}_i}{\partial \pival_k} &\eqdef \frac{\partial}{\partial \pival_k} \sum_{\substack{(a, b) \in E\\i = a}}q_{ib} \cdot \hat{V}_b(\pival)\\
      &=\Pr(i \rightsquigarrow j \mid \pival) \cdot \begin{cases}
        0 & \text{if } \Pr(i \rightsquigarrow k) = 0\\
        \frac{\partial}{\partial \pival_k} \hat{V}_j(\pival) &
        \text{otherwise},
      \end{cases}
    \end{split}
  \end{equation}
  Similarly, note that because the derivative of logsumexp is the
  softmax function, for any ego node $i$,
  \begin{equation}\label{eq:lem2}
    \begin{split}
      \frac{\partial \hat{V}_i}{\partial \pival_k} &\eqdef \frac{\partial}{\partial \pival_k} \log\sum_{\substack{(a, b) \in E\\i = a}}\hat{Q}_b(\pival)\\
      &=\begin{cases}
        0 & \text{if } \Pr(i \rightsquigarrow k) = 0\\
        e^{\hat{Q}_j(\pival) - \hat{V}_i(\pival)}\cdot
        \frac{\partial}{\partial \pival_k} \hat{Q}_j(\pival) &
        \text{otherwise},
      \end{cases}
    \end{split}
  \end{equation}
  where again, $j$ denotes the (potential) unique child of $i$ that
  can reach $k$. Finally, observe that by definition
  $e^{\hat{Q}_j(\pival) - \hat{V}_i(\pival)} = \Pr(i \rightsquigarrow
  j \mid \pival)$, using the maximum entropy policy induced by
  $\pival$. 
  Substituting into~\eqref{lem1}, we see that the lemma
  follows by induction
\end{proof}

\begin{proof}[Prop~\ref{prop:grad}]
  Recall that the probability of traversing an environment edge is
  constant w.r.t $\pival$. Thus, inspecting~\eqref{proxy} we see that
  it suffices to prove that for any ego edge, $(i, j)$,
  \[
  \frac{\partial}{\partial\pival_k} \ln\Pr(i\rightsquigarrow j \mid
  \pival)= \Pr(i \rightsquigarrow k\mid \pival) - \Pr(j
  \rightsquigarrow k \mid \pival).\] Recall that by definition, if $i$
  is ego, then
  $\ln\Pr(i\rightsquigarrow j \mid \pival)= \hat{Q}_i(\pival) -
  \hat{V}_i(\pival)$. Thus, the proposition follows directly from
  Lemma~\ref{lem:grad}.
\end{proof}
\section{DISS Pseudo Code}\label{sec:diss-pseudo}
\begin{algorithm}[h]
  \caption{Demonstration Informed Specification
    Search.\label{alg:diss}}
  \begin{algorithmic}[1]
    \State \textbf{input: }
    $(\trc_1, \ldots, \trc_m), M, \theta, N, \kappa$
    \State Compute $\prefixtree$ given $(\trc_1, \ldots, \trc_m)$.
    \hfill\Comment{Create prefix tree.}
    \State $\RepresentationClass \gets \emptyset$.
    \For{$t$ in $1,\ldots, N$}
      \If{$t \equiv 0 \pmod \kappa$}
        \State $\EX \sim \argmax_{\psi \in \RepresentationClass} U(\psi)$
        \hfill\Comment{Reset periodically.}
        \State $d\EX \gets \emptyset$ \EndIf
      \State $\EX' \gets \text{update}(\EX', d\EX)$
      \hfill\Comment{Newer label wins under conflict.}
      \State $\spec' \sim \identify(\EX')$.  \hfill\Comment{Sample
        candidate task.}
      \State $\RepresentationClass \gets \RepresentationClass \cup \{\spec'\}$.
      \hfill\Comment{Update visited specs.}
      \State $\temperature \gets \text{cooling\_schedule}(t)$
      \hfill\Comment{User defined.}
      \State $dU \gets U(\spec') - U(\spec)$
      \State $\alpha \sim \text{Uniform}(0, 1)$
      \If{$dU < 0$ or $\exp(-\nicefrac{dU}{T}) \leq \alpha$}
        \State $(\spec, \EX) \gets (\spec', \EX')$
        \State $d\EX \gets \{\text{SGS}(\spec, T, M)\}$
        \hfill\Comment{Conjecture labeled example.}  \Else
        \State $d\EX \gets \emptyset$ \hfill\Comment{Reject proposal.}
      \EndIf \EndFor
    \State \textbf{return} $\RepresentationClass$
  \end{algorithmic}
\end{algorithm}

\end{document}